\documentclass[prl,twocolumn,twoside,showpacs,byrevtex,superscriptaddress]{revtex4}

\usepackage{currfile}

\usepackage{graphicx,epsfig,verbatim,enumerate}
\usepackage{amssymb,amsmath}
\usepackage{ifthen}

\usepackage{tikz-cd}
\usetikzlibrary{arrows}
\tikzset{
  commutative diagrams/.cd,
  arrow style=tikz,
  diagrams={>=space}}

\newboolean{twocolswitch}

\newcommand{\sindex}[1]{}
\newcommand{\nindex}[1]{}

\newcommand{\www}[1]{\url{#1}}

\usepackage{lettrine}

\usepackage{color}

\setboolean{twocolswitch}{true}

\begin{document}

\title{
  Zipf's law is a consequence of coherent language production

}

\author{
\firstname{Jake Ryland}
\surname{Williams}
}

\email{jakerylandwilliams@berkeley.edu}

\affiliation{School of Information,
  University of California, Berkeley
  102 South Hall \#4600
  Berkeley, CA 94720-4600.}

\author{
\firstname{James P.}
\surname{Bagrow}
}

\email{james.bagrow@uvm.edu}

\affiliation{Department of Mathematics \& Statistics,
  Vermont Complex Systems Center,
  Computational Story Lab,
  \& the Vermont Advanced Computing Core,
  The University of Vermont,
  Burlington, VT 05401.}

\author{
\firstname{Andrew J.}
\surname{Reagan}
}

\email{Andrew.Reagan@uvm.edu}

\affiliation{Department of Mathematics \& Statistics,
  Vermont Complex Systems Center,
  Computational Story Lab,
  \& the Vermont Advanced Computing Core,
  The University of Vermont,
  Burlington, VT 05401.}

\author{
\firstname{Sharon E.}
\surname{Alajajian}
}

\email{sharon@ischool.berkeley.edu}

\affiliation{School of Information,
  University of California, Berkeley
  102 South Hall \#4600
  Berkeley, CA 94720-4600.}

\author{
\firstname{Christopher M.}
\surname{Danforth}
}
\email{chris.danforth@uvm.edu}

\affiliation{Department of Mathematics \& Statistics,
  Vermont Complex Systems Center,
  Computational Story Lab,
  \& the Vermont Advanced Computing Core,
  The University of Vermont,
  Burlington, VT 05401.}

\author{
\firstname{Peter Sheridan}
\surname{Dodds}
}
\email{peter.dodds@uvm.edu}

\affiliation{Department of Mathematics \& Statistics,
  Vermont Complex Systems Center,
  Computational Story Lab,
  \& the Vermont Advanced Computing Core,
  The University of Vermont,
  Burlington, VT 05401.}

\date{\today}

\begin{abstract}
  The task of text segmentation
may be undertaken at many levels in text
analysis---paragraphs, sentences, words, or even letters.
Here, we focus on a relatively fine scale of segmentation,
hypothesizing it to be in accord with a stochastic model of language generation,
as the smallest scale where independent units of meaning are produced.
Our goals in this letter include the development of methods for
the segmentation of these minimal independent units,
which produce feature-representations of texts
that align with the independence assumption of the bag-of-terms model,
commonly used for prediction and classification in computational text analysis.
We also propose the measurement of texts' association
(with respect to realized segmentations)
to the model of language generation.
We find (1) that our segmentations of phrases
exhibit much better associations to the generation model than words
and (2), that texts which are well fit are generally topically homogeneous.
Because our generative model produces Zipf's law,
our study further suggests that
Zipf's law may be a consequence of homogeneity in language production.
  
\end{abstract}


\maketitle

In our previous work on text partitioning~\cite{williams2014a},
and again in our study on the effects of text mixing~\cite{williams2014b},
we have observed the relevance of the selection model
proposed long ago by Simon~\cite{simon1955a}.
We will build on this model again,
together with Zipf's law~\cite{zipf1935a,zipf1949a},
as the basis for a model of frequency data of text.

The defining aspects of Simon's model for language generation
hold that as the terms of a text appear,
they do so independently,
and in proportion to their historic frequencies of occurrence.
The first assumption (of memoryless term-term independence) is
likewise the basis for the bag-of-terms model
that is pervasive in the computational text analysis community.
It is evident that this assumption fails for strictly defined words
when one considers the dependence of terms
that are bound, like {\it New}, {\it York}, and {\it City}.
As we have argued in~\cite{williams2014b},
dependence can also be caused by mixing,
where the rate of a term's occurrence
varies in a composite corpus of many texts
as one transitions from document to document.

The two types of dependence indicate that
for the Simon model to apply,
and consequently,
for the bag-of-terms model to apply,
texts should be analyzed on the homogeneous (unmixed) scales of production
(i.e., by writer, or even topic),
and that the terms of texts should be segmented
according to their local dependence
(i.e., grouped together into irreducible expressions),
rendering distributions of terms of mixed sizes.
The first criterion can be relatively easy to satisfy,
and in our analyses will come for free
by the curation of the Project Gutenberg eBooks~\cite{gutenberg2010} database.
The second, however, poses a much greater challenge,
as the meanings that bind words into meaning-irreducible forms
are only known a priori to the writers who generate text for their meanings.

In our work on stochastic text partitioning~\cite{williams2014a},
we proposed a simple mechanism for the fine-grained
chunking of text into phrases.
Under this framework,
sentences were broken down stochastically,
splitting on whitespace with a
uniform bond-breakage probability, $q$.
While this (uniform) framework is clearly
not the ideal mechanism to isolate phrases
for, say, dictionary lookups,
we did find that random partitioning of phrases ($q=\frac{1}{2}$)
produced Zipf laws extending many orders of magnitude in rank
beyond the standard version for words only.

Here, we wish to explore informed partitions,
taking into account empirical segmentation information.
Some interesting options for estimating bond-breakage frequencies exist---in
language learning environments,
similar data are generated through exercises referred to as phrase-cued text,
which have been shown to help learners develop prosody~\cite{glavach2011a}
(the patterns of stress and intonation,
considered to be one of the essential features of reading fluency~\cite{miller2006a}).
This suggests value in the phrase-scale of segmentation.

To obtain phrases, surveys (essentially classroom exercises) could be run
that would benefit educators and students,
while developing a base of training data for informed partitions.
Ideally, such surveys would have students learning a language
asked to segment their \emph{own} writing.
However, this is unconventional in the classroom,
and poses a more difficult task to request.
So, for now we will turn to expertly-segmented data.

In the computational linguistics (CL) community
there have been recent efforts to establish benchmarks
and testing sets for multiword expressions (MWEs)
(loosely defined as groups of
``tokens in a sentence that cohere
more strongly than ordinary syntactic combinations'')
extraction~\cite{schneider2014a}.
While these data can be used to inform a partitioner,
they are alone insufficient (narrowly focused on business reviews),
and are better used for testing and benchmarking.
Furthermore, it is not clear if MWEs and their extraction
(used for work in information retrieval and other CL tasks)
coincide with language generation and distributional representations,
which are the focus, here.
We have also executed a companion piece to this study~\cite{williams2016b},
focused on the task of MWE extraction,
where the text partitioning framework was tuned
to exhibit its ability to perform as strongly,
if not better than, the state-of-the-art~\cite{schneider2015a},
with very substantial increases in speed,
while being trained on less data.

For training, we turn to the bond-breakage information
held latently inside of the Wiktionary~\cite{wiktionary2014}
with its trove of larger-than-word entries
and their example usages.
Noting that example usages of phrases in the Wiktionary
are consistently coded in boldfaced text,
we are able to resolve isolated empirical bond-breakage frequency data.
For example, at the time of writing this letter,
the entry for the phrase
``have a ball''
was represented by the usage:
\begin{center}
The kids {\bf had a ball} playing in the fountain.
\end{center}
which informed us that $(\text{kids},\text{had})$
is a pair more likely split,
and that $(\text{had},\text{a})$
is a pair more likely bound.
The Wiktionary examples have the added bonus of being trustworthy,
on account of their being archetypal-usage examples,
and will serve to inform our partitions for all of the experiments in this letter.
However, to make informed text partitions available
in many other languages
(where examples may be less prolific and/or coded differently),
we also construct bond-breakage frequencies in a similar manner,
but from the embedded hyperlinks on Wikipedia,
which are ubiquitous and coded-for consistently across languages.
From either source of data,
we define the partitioning probability,
$q(w_\ell,w_\text{r})$,
for an ordered pair of words, $(w_\ell,w_\text{r})$,
according to the total number of known bond breaking appearances
$f_\text{b}(w_\ell,w_\text{r})$
and the total number of bond preserving appearances
$f_\text{p}(w_\ell,w_\text{r})$:
$$q(w_\ell,w_\text{r}) = \frac{f_\text{b}(w_\ell,w_\text{r})}{f_\text{p}(w_\ell,w_\text{r}) +f_\text{b}(w_\ell,w_\text{r})},$$
with a default value of $q(w_\ell,w_\text{r}) = 1$
for all pairs unobserved in a training set.
Note that this model is still quite na\"{i}ve,
and may be improved by allowing values of $q$ to vary,
depending on terms existing farther away in a text.
However, training a more sophisticated partitioner like this
would be done best if the training data were fully segmented
(which might be accomplished with phrase-cued text).

We note that both sources (examples and hyperlinks)
have a reasonable degree of accord (considering English),
with the resulting $q$-values
falling on the same side of $q=0.5$ for $\sim 65\%$ of all unique word pairs
(which later will be all that is necessary to ensure equivalent partitions),
and as much as $75\%$ for those most frequently appearing.
While the two sources share coverage on about $20,000$ word pairs
(and disagree on a substantial number),
we note that the Wikipedia hyperlinks source is much larger,
with coverage on many proper nouns (names of places, people, etc.),
which were found to be important for MWE identification in~\cite{williams2016b}.

In our original work on text partitioning~\cite{williams2014a},
we considered stochastic text partitions,
where bonds between pairs words
would either be preserved or broken at random (uniformly).
Despite our solving for the analytic expectation across all possible partitions
of phrase-partition frequencies,
it became necessary to execute one-off random partitions
for the study of true rank-frequency distributions.
However for CL tasks, these stochastic partitions have little value
on account of their variation, and while the variation in the case
of informed stochastic partitions is reduced,
it is still not ideal, as CL tasks require consistency.

A perhaps unexpected bonus to the na\"{i}vet\'{e} of our assumptions
(specifically, defining the q's to be independent of one another)
is that the maximum likelihood partition (MLP),
or most frequently occurring partition of a text,
is easily computed.
Accepting this partition (for a text) is deterministic,
and accomplished simply by breaking a bond
whenever its bounding pair of words have a breakage probability
of at least $0.5$---note that this conservatively
sets the ill-determined case of $q = 0.5$ to bond-breaking.
Using this scheme, a text can be segmented very rapidly
based on the Wiktionary
(with the same computational complexity as simple word count),
and requires no global information.
For all of these reasons---speed, determinism,
and maximum likelihood---we continue with the MLP in this work.

We now return to Simon's model of language generation,
which analytically produces single-parameter power-law
rank-frequency distributions,
and is our hypothesis as the dominant
productive mechanism for Zipf's law.
Zipf's law succinctly formulates a relationship
between the frequencies of occurrence of terms,
and their ranks by frequency (descending):
\begin{equation}
  f = C\cdot r^{-\theta},
\end{equation}
for a scaling constant, $C$.
Simon's model has one tunable parameter---the rate of term introduction, $\alpha$.
When language is generated by Simon's model,
Zipf's scaling exponent is given by $\theta = 1 - \alpha$,
and in the case of an empirical text,
the term introduction rate can be inferred as $\alpha = N/M$,
where $N$ is the number of unique term types in the text,
and $M$ is the total number of all terms occurring in a text.

The main caveat for the compatibility of Zipf's empirical law
with Simon's theoretical model
arises from the fact that Simon's model is only able to generate 
single-scaling rank-frequency distributions
with scaling parameters $\theta$ in $[0,1)$.
While many texts exhibit
multiple scalings and values of $\theta$ larger than $1$,
we have seen that these anomalies are often
the result of text mixing and term dependence~\cite{williams2014b}.
Consequently, we view the degree to which
a particular Zipf/Simon model aligns with
empirical rank-frequency data
as a measure of the
goodness of fit for the `bag-of-terms'
representation (tokenization) of a text
(regardless of how the terms are defined, i.e., words or phrases).

\begin{figure}[t!]
  \includegraphics[width=0.5\textwidth, angle=0]{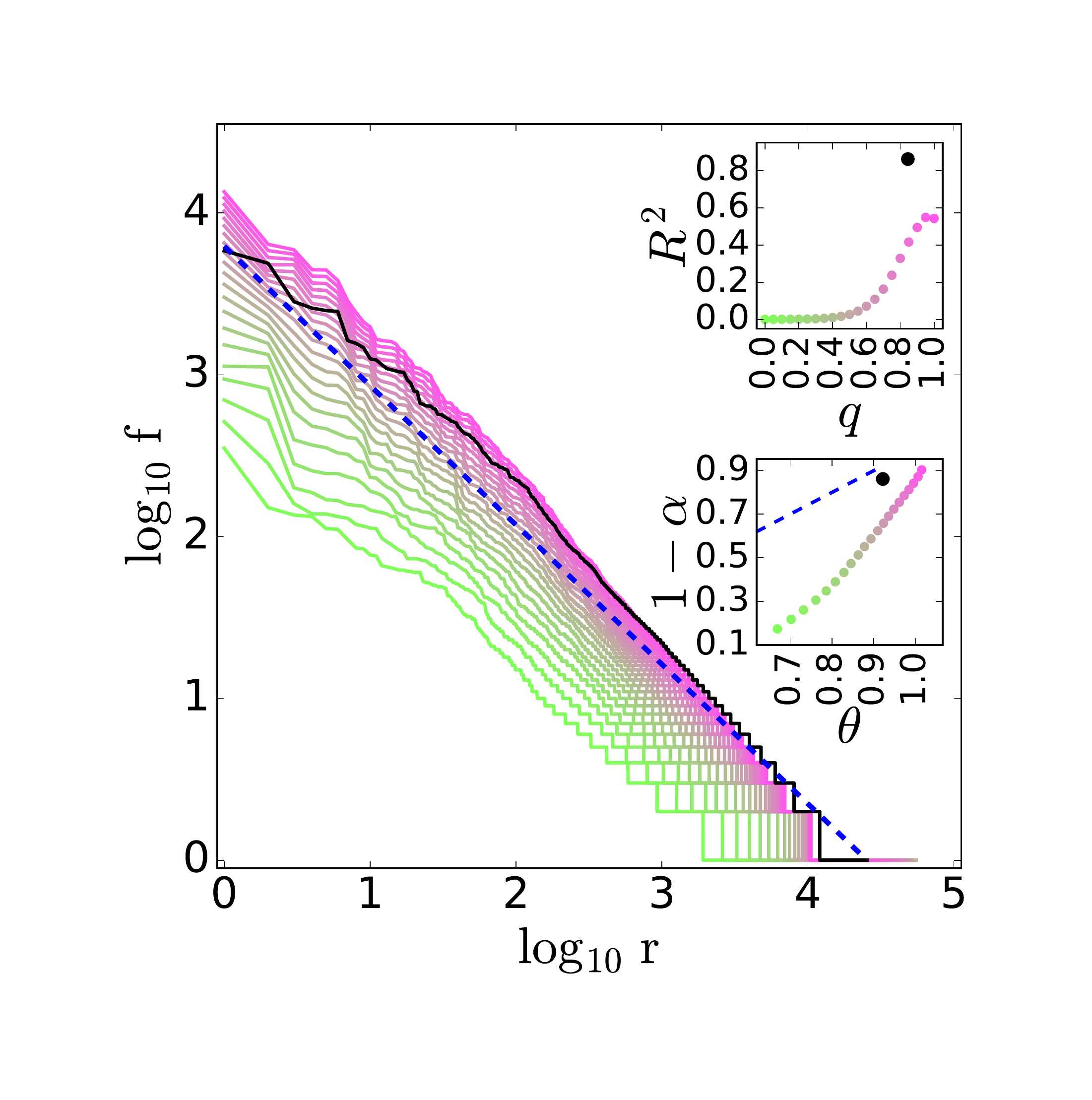}
  \caption{
    The main axes show rank-frequency distributions
    from Herman Melville's ``Moby Dick''
    for one-off uniform stochastic partitions
    for values of $q$ ranging from $0$ (green) to $1$ (pink),
    together with a one-off, dictionary-informed stochastic partition (black)
    and its Zipf/Simon model (blue, dashed).
    In the upper inset we present
    the Zipf/Simon goodness of fit
    as a function of $q$
    (with the black point for the informed partition
    positioned at the text's average value of $q$),
    and in the lower inset we present
    regressed scalings ($\theta$)
    against those determined by the Zipf/Simon model ($1 - \alpha$).
    In the lower inset,
    complete agreement occurs along the (blue dashed)
    line $1 - \alpha = \theta$,
    and for both insets,
    colors of points correspond to
    those in the main axes.
  }
  \label{fig:mobyDists}
\end{figure}

The Zipf/Simon fit for a text of $N$ unique and $M$ total words is defined as follows.
Assuming a text was generated precisely according to Simon's model,
the constant word introduction rate is inferred by the text-wide average: $\alpha = N/M$,
and an estimate of the scaling exponent is then $\theta_\text{mod} = 1 - \alpha$.
The exact form of the model's fit is then obtained by computing the constant of proportionality:
$C_\text{mod} = N^{\theta_\text{mod}}$,
whereupon we may take the coefficient of determination, or $R^2$,
as a measure of goodness of fit for the Zipf/Simon model,
and as a result of the Simon model's independence assumption,
a measure of the quality of a text's bag-of-terms representation.
Note that this will also give us an opportunity to evaluate
the qualitative nature of language produced by Simon's model.

Since the connection of Simon's model to the parameter $\theta$
occurs naturally in the complimentary cumulative distribution function (CCDF)
of term frequencies,
we measure and regress all quantities
along CCDFs,
while we present all results
in the intuitive and familiar
rank-frequency (Zipf) representations.

In Fig.~\ref{fig:mobyDists},
we plot an example informed partition
for Herman Melville's ``Moby Dick''
(black line, main axes), and behind it,
the spectrum of uniform stochastic partitions
ranging from $q=0$ (green) sentences/clauses,
to $q=1$ (pink) words.
One can see the distributions steepen along the gradient
as $q$ is increased
(which is generally the trend throughout the eBooks),
and as much can be seen in the lower inset,
where a regressed scaling parameter $\theta$
is compared with $\theta_\text{mod} = 1 - \alpha$.
Also within the main axes lies the Zipf/Simon fit
for the the informed partition (blue, dashed),
which appears reasonable by inspection,
and outperforms all uniform partitions
(including the $q=1$ word partition)
by goodness of fit of the Zipf/Simon model
(upper inset),
where the informed partition
is indicated by the black point (in both insets).
Note that this increased quality of fit
by the Zipf/Simon model is also 
observed readily in the lower inset,
where $\theta$ and $1-\alpha$ nearly align.

\begin{figure}[t!]
  \includegraphics[width=0.5\textwidth, angle=0]{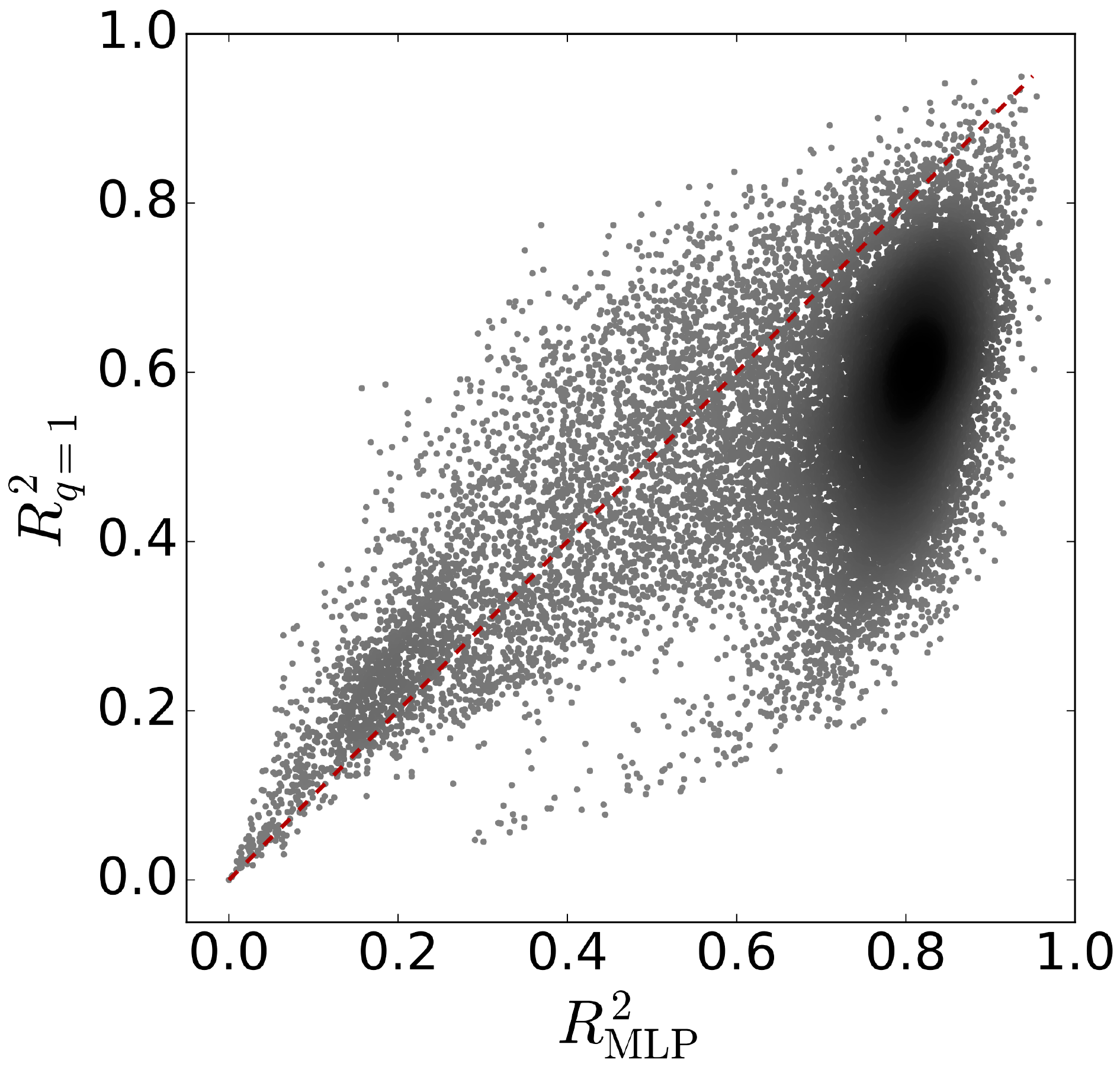}
  \caption{
    We plot the goodness of fit ($R^2$) of the Zipf/Simon model,
    applied to the texts of the English Project Gutenberg eBooks database,
    where the dictionary-informed one-off partitions (horizontal axis)
    are plotted against the $q=1$ word partitions (vertical axis).
    The discriminating line (red, dashed, $R^2_\text{MLP} = R^2_{q = 1}$)
    helps divide the collection into texts
    that are word-based and phrase-based.
  }
  \label{fig:allBooks}
\end{figure}

\begin{figure}[t!]
  \includegraphics[width=0.5\textwidth, angle=0]{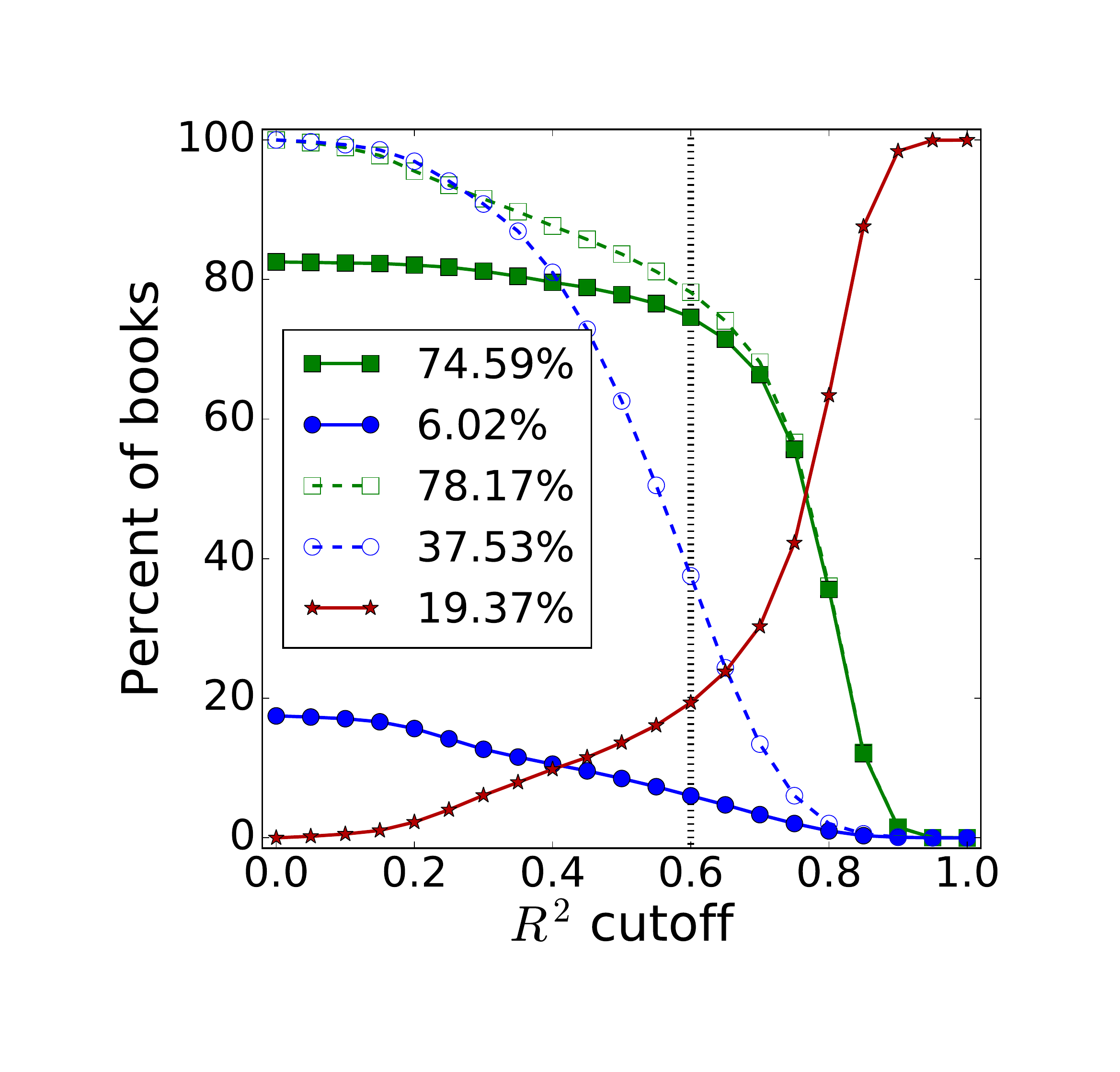}
  \caption{
    The percent of books
    in the Project Gutenberg English eBooks database
    that are phrase-based (solid green line, filled squares),
    word-based (solid blue line, filled circles),
    and poorly fit (red line, stars),
    when a cutoff in $R^2$ is applied (right).
    A dropoff in $R^2_\text{MLP}$ occurs near $R^2 \sim 0.6$,
    which is denoted by the vertical, black dotted line.
    We also present curves that indicate the percents of books
    remaining above the cutoff when blanketed usage of either
    words (blue dashed line, open circles) or
    informed phrases (green dashed line, open squares) are applied.
  }
  \label{fig:numBooks}
\end{figure}

Zooming out to the larger collection of (roughly $20,000$) English eBooks,
we focus on comparing two partitions for each text---the
$q=1$ (word) partitions, and the informed (phrase) partitions.
In Fig.~\ref{fig:allBooks},
we compare the goodness of fit by the Zipf/Simon model
between the two partition types.
This comparison divides the collection of books
into two subsets that we will refer to as
the word-based and phrase-based texts.
Our demarcation is given by the line
$R^2_{q=1} = R^2_\text{MLP}$ (red, dashed),
and shows that most texts
were better bag-models under the informed
(phrase-based, $R^2_{q=1} < R^2_\text{MLP}$)
partition framework.
While a good number of texts ($\sim 20\%$) are word-based,
we note that these texts tend to have weaker $R^2$ values in general,
which can be seen
in Fig.~\ref{fig:numBooks}
where we apply an $R^2$ cutoff to the books
and measure the portion of the data set removed.
This quantifies the numbers of phrase-and word-based texts
that are strong fits ($R^2 > 0.6$) to be in a proportion of more than $12$:$1$, respectively,
indicating that a strongly-fit bag-of-terms representation of a text
is better (and generally quite significantly) achieved by MLP phrases
more than $90\%$ of the time.

From Fig.~\ref{fig:numBooks},
we can see that $R^2\sim0.6$
appears near a sharp drop for the cutoff, which,
if applied as a threshold for analysis,
results in approximately a $20\%$ loss (red line/stars) for the dataset
(largely due to the low-$R^2$, word-based texts).
By allowing the books to be either phrase- or word-based
(according to their $R^2$ values),
we are then able to accommodate a strong ($R^2 \geq 0.6$) `bag' analysis
for over $80\%$ of the dataset,
which is a large improvement over the conventional,
blanket usage of words (dashed blue line/open circles),
which only accommodates strong fits
for approximately $37.5\%$ of the collection.

While one can view the $\sim 20\%$ of poorly-fit
books as a loss to quality analysis,
it is possible that refinement of the partition function
(through better data or modeling)
would allow one to obtain better tokenizations/fits
for all books.
However, one can also look at these $20\%$ in a different way---as
potentially being unfit for the bag-of-terms framework.
Sorting the eBooks according to $\max\{R^2_{q=1},R^2_\text{MLP}\}$,
we find the ten poorest fits
to include dictionaries, spelling books, and
books of extremely small size
(often as placeholders for other media),
indicating that low $R^2$ values
appropriately identify texts unfit for analyses.

\begin{figure}[t!]
  \includegraphics[width=0.5\textwidth, angle=0]{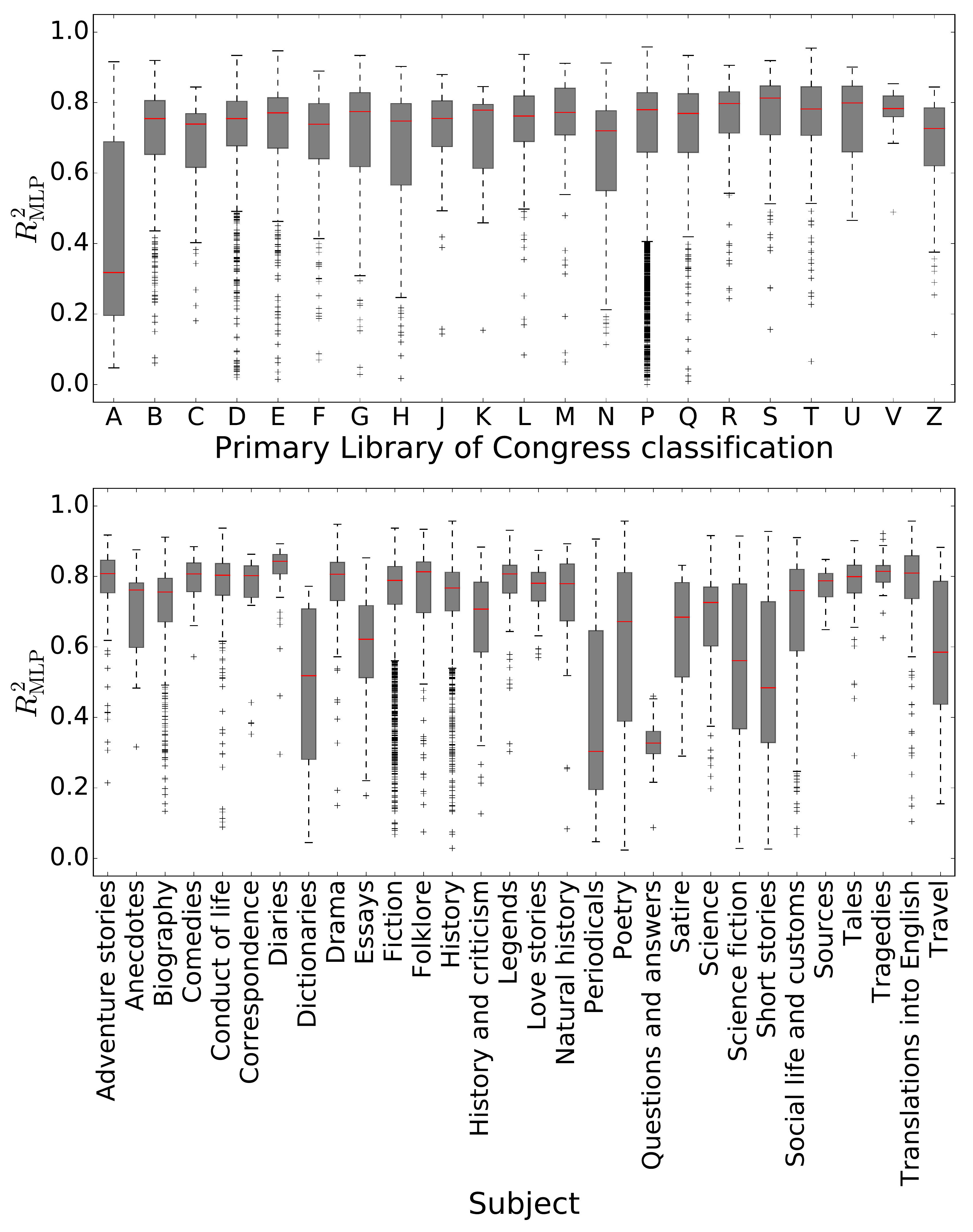}
  \caption{
    Box plots, showing variation of $R^2_\text{MLP}$ across
    the primary Library of Congress classifications (top),
    and a selection of the Project Gutenberg eBooks subjects (bottom).
    All boxes represent at least 50 texts.
  }
  \label{fig:boxplots}
\end{figure}

To explore the qualitative nature of fit quality and the Zipf/Simon model,
we also consider the variation of $R^2_\text{MLP}$
(strictly, as $R^2_\text{q=1}$ is so frequently outperformed)
across the different Library of Congress classifications (Fig.~\ref{fig:boxplots}, top),
and subjects (Fig.~\ref{fig:boxplots}, bottom) provided in the eBooks collection.
This variation can be seen in Fig.~\ref{fig:boxplots},
where we observe stark differences in the ranges of box plots.
As the majority of texts have an MLP representation 
that is relatively strong ($R^2_\text{MLP} > 0.6$),
the medians for most classifications are high.
However, it is clear that texts under classification A---general works,
which includes dictionaries, encyclopedias, and periodicals---are more often poorly-fit.
This is of note, as all other classifications,
including textbook-reference classifications,
such as medicine (R), law (K), and naval (V) and military (U) science,
exhibit medians substantially above $0.6$,
though we note that many classifications exhibit
outliers of low $R^2_\text{MLP}$,
notably with language and literature (P),
which contains long, brief, and mixed texts.

The main difference we note between classification A
and the other reference classifications mentioned
is that the general works of A are
frequently not topically homogeneous.
This observation is echoed in the bottom of Fig.~\ref{fig:boxplots},
where we consider the eBooks' subjects.
Here, we again see low values for dictionaries and periodicals,
but likewise for other heterogeneous subjects,
such as questions and answers, travel, and social life.
We also find that poetry, short stories, and science fiction
all exhibit a large proportion of low $R^2_\text{MLP}$ values.
These subjects all share the commonality of frequent brevity.
This makes some sense in relation to the Simon model,
as the generation of a text may not
have an opportunity to stabilize (distributionally).
We also note that the science fiction subject
may frequently use phrases that are
outside of the informed partitioner's training.

Our results show that informed text partitioning
is capable of improving the basic methodologies
of feature selection in text analysis on a broad scale,
allowing for better adherence to assumptions
(specifically, for the bag-of-terms model)
that underpin a vast collection of
algorithms currently in practice
throughout industry and academia.
Additionally, phrase-based text analysis
improves the independent interpretability of features---in the soft sense,
at the level of user experience.
With phrase-based text analysis,
end-point users
(e.g., policy makers or product users interpreting
lists of phrase-feature topics from a topic model readout)
who may not understand the workings of an algorithm
will be better able to interpret results,
as phrases provide critical context for interpretation.
We have also
proposed the measurement of association to
a model of language generation,
which we have found here to be the likely mechanism
for the production of topically-homogeneous language,
allowing us to assert new meaning
for the presence of Zipf's law in natural language---long observed,
though largely a mystery.
We also suggest the possibility
that as phrase-segmentation methods improve,
association to the Zipf/Simon model
might also be used as a metric for determination
of topical breaks in text,
since $R^2_\text{MLP}$ does appear to increase
in the presence of topically-homogeneous texts.

To make these tools both explorable and available
to the computational text analysis community,
we have with this letter developed
a Python package for text partitioning
(for detailed information, see \href{https://github.com/jakerylandwilliams/partitioner}{https://github.com/jakerylandwilliams/partitioner}),
which makes our tools available in 11 languages
(English, German, Russian, Portuguese, Polish, Dutch,
Italian, French, Finnish, Spanish and Greek),
and in addition present an explorable online appendix
(\href{http://jakerylandwilliams.github.io/partitioner/}{http://jakerylandwilliams.github.io/partitioner/}).


\begin{thebibliography}{12}
\expandafter\ifx\csname natexlab\endcsname\relax\def\natexlab#1{#1}\fi
\expandafter\ifx\csname bibnamefont\endcsname\relax
  \def\bibnamefont#1{#1}\fi
\expandafter\ifx\csname bibfnamefont\endcsname\relax
  \def\bibfnamefont#1{#1}\fi
\expandafter\ifx\csname citenamefont\endcsname\relax
  \def\citenamefont#1{#1}\fi
\expandafter\ifx\csname url\endcsname\relax
  \def\url#1{\texttt{#1}}\fi
\expandafter\ifx\csname urlprefix\endcsname\relax\def\urlprefix{URL }\fi
\providecommand{\bibinfo}[2]{#2}
\providecommand{\eprint}[2][]{\url{#2}}

\bibitem[{\citenamefont{Williams
  et~al.}(2015{\natexlab{a}})\citenamefont{Williams, Lessard, Desu, Clark,
  Bagrow, Danforth, and Dodds}}]{williams2014a}
\bibinfo{author}{\bibfnamefont{J.~R.} \bibnamefont{Williams}},
  \bibinfo{author}{\bibfnamefont{P.~R.} \bibnamefont{Lessard}},
  \bibinfo{author}{\bibfnamefont{S.}~\bibnamefont{Desu}},
  \bibinfo{author}{\bibfnamefont{E.~M.} \bibnamefont{Clark}},
  \bibinfo{author}{\bibfnamefont{J.~P.} \bibnamefont{Bagrow}},
  \bibinfo{author}{\bibfnamefont{C.~M.} \bibnamefont{Danforth}},
  \bibnamefont{and} \bibinfo{author}{\bibfnamefont{P.~S.} \bibnamefont{Dodds}},
  \bibinfo{journal}{CoRR} \textbf{\bibinfo{volume}{abs/1406.5181}}
  (\bibinfo{year}{2015}{\natexlab{a}}),
  \bibinfo{note}{\href{http://arxiv.org/abs/1406.5181}{http://arxiv.org/abs/1406.5181}}.

\bibitem[{\citenamefont{Williams
  et~al.}(2015{\natexlab{b}})\citenamefont{Williams, Bagrow, Danforth, and
  Dodds}}]{williams2014b}
\bibinfo{author}{\bibfnamefont{J.~R.} \bibnamefont{Williams}},
  \bibinfo{author}{\bibfnamefont{J.~P.} \bibnamefont{Bagrow}},
  \bibinfo{author}{\bibfnamefont{C.~M.} \bibnamefont{Danforth}},
  \bibnamefont{and} \bibinfo{author}{\bibfnamefont{P.~S.} \bibnamefont{Dodds}},
  \bibinfo{journal}{CoRR}  (\bibinfo{year}{2015}{\natexlab{b}}),
  \bibinfo{note}{\href{http://arxiv.org/abs/1409.3870}{http://arxiv.org/abs/1409.3870}}.

\bibitem[{\citenamefont{Simon}(1955)}]{simon1955a}
\bibinfo{author}{\bibfnamefont{H.~A.} \bibnamefont{Simon}},
  \bibinfo{journal}{Biometrika} \textbf{\bibinfo{volume}{42}},
  \bibinfo{pages}{425} (\bibinfo{year}{1955}).

\bibitem[{\citenamefont{Zipf}(1935)}]{zipf1935a}
\bibinfo{author}{\bibfnamefont{G.~K.} \bibnamefont{Zipf}},
  \emph{\bibinfo{title}{The Psycho-Biology of Language}}
  (\bibinfo{publisher}{Houghton-Mifflin}, \bibinfo{year}{1935}).

\bibitem[{\citenamefont{Zipf}(1949)}]{zipf1949a}
\bibinfo{author}{\bibfnamefont{G.~K.} \bibnamefont{Zipf}},
  \emph{\bibinfo{title}{Human Behaviour and the Principle of Least-Effort}}
  (\bibinfo{publisher}{Addison-Wesley}, \bibinfo{year}{1949}).

\bibitem[{\citenamefont{Project{~}Gutenberg}(2010)}]{gutenberg2010}
\bibinfo{author}{\bibnamefont{Project{~}Gutenberg}} (\bibinfo{year}{2010}),
  \bibinfo{note}{\href{http://www.gutenberg.org}{http://www.gutenberg.org}}.

\bibitem[{\citenamefont{Glavach}(2011)}]{glavach2011a}
\bibinfo{author}{\bibfnamefont{M.~J.} \bibnamefont{Glavach}},
  \bibinfo{journal}{The Practical Teacher}  (\bibinfo{year}{2011}).

\bibitem[{\citenamefont{Miller and Schwanenflugel}(2006)}]{miller2006a}
\bibinfo{author}{\bibfnamefont{J.}~\bibnamefont{Miller}} \bibnamefont{and}
  \bibinfo{author}{\bibfnamefont{P.~J.} \bibnamefont{Schwanenflugel}},
  \bibinfo{journal}{Journal of Educational Psychology}
  \textbf{\bibinfo{volume}{98}}, \bibinfo{pages}{839} (\bibinfo{year}{2006}).

\bibitem[{\citenamefont{Schneider et~al.}(2014)\citenamefont{Schneider,
  Onuffer, Kazour, Danchik, Mordowanec, Conrad, and Smith}}]{schneider2014a}
\bibinfo{author}{\bibfnamefont{N.}~\bibnamefont{Schneider}},
  \bibinfo{author}{\bibfnamefont{S.}~\bibnamefont{Onuffer}},
  \bibinfo{author}{\bibfnamefont{N.}~\bibnamefont{Kazour}},
  \bibinfo{author}{\bibfnamefont{E.}~\bibnamefont{Danchik}},
  \bibinfo{author}{\bibfnamefont{M.~T.} \bibnamefont{Mordowanec}},
  \bibinfo{author}{\bibfnamefont{H.}~\bibnamefont{Conrad}}, \bibnamefont{and}
  \bibinfo{author}{\bibfnamefont{N.~A.} \bibnamefont{Smith}}, in
  \emph{\bibinfo{booktitle}{Proceedings of the Ninth International Conference
  on Language Resources and Evaluation (LREC'14)}}
  (\bibinfo{publisher}{European Language Resources Association (ELRA)},
  \bibinfo{address}{Reykjavik, Iceland}, \bibinfo{year}{2014}), ISBN
  \bibinfo{isbn}{978-2-9517408-8-4}.

\bibitem[{\citenamefont{Williams}(2016)}]{williams2016b}
\bibinfo{author}{\bibfnamefont{J.~R.} \bibnamefont{Williams}},
  \bibinfo{journal}{CoRR}  (\bibinfo{year}{2016}),
  \urlprefix\url{http://people.ischool.berkeley.edu/~jakeryland/documents/williams2016b.pdf}.

\bibitem[{\citenamefont{Schneider and Smith}(2015)}]{schneider2015a}
\bibinfo{author}{\bibfnamefont{N.}~\bibnamefont{Schneider}} \bibnamefont{and}
  \bibinfo{author}{\bibfnamefont{N.~A.} \bibnamefont{Smith}}, in
  \emph{\bibinfo{booktitle}{Proceedings of the 2015 Conference of the North
  American Chapter of the Association for Computational Linguistics: Human
  Language Technologies}} (\bibinfo{publisher}{Association for Computational
  Linguistics}, \bibinfo{address}{Denver, Colorado}, \bibinfo{year}{2015}), pp.
  \bibinfo{pages}{1537--1547},
  \urlprefix\url{http://www.aclweb.org/anthology/N15-1177}.

\bibitem[{\citenamefont{Wiktionary}(2014)}]{wiktionary2014}
\bibinfo{author}{\bibnamefont{Wiktionary}} (\bibinfo{year}{2014}),
  \bibinfo{note}{\href{http://dumps.wikimedia.org/enwiktionary/}{http://dumps.wikimedia.org/enwiktionary/}2014}.

\end{thebibliography}
\end{document}